# A Novel Approach For Generating Customizable Light Field Datasets for Machine Learning

Julia Huang
Northville High School
SHTEM Program
Stanford University
Northville, Michigan
juliah6169@gmail.com

Toure Smith
Abraham Lincoln HS
SHTEM Program
Stanford University
San Jose, California
touresmith10@gmail.com

Aloukika Patro
Douglas County HS
SHTEM Program
Stanford University
Castle Rock, Colorado
patro.aloukika@gmail.com

Vidhi Chhabra
West Haven High School
SHTEM Program
Stanford University
West Haven, Connecticut
vidhichhabra06@gmail.com

*Abstract*—To train deep learning models, which often outperform traditional approaches, large datasets of a specified medium, e.g., images, are used in numerous areas. However, for light field-specific machine learning tasks, there is a lack of such available datasets. Therefore, we create our own light field datasets, which have great potential for a variety of applications due to the abundance of information in light fields compared to singular images. Using the Unity and C# frameworks, we develop a novel approach for generating large, scalable, and reproducible light field datasets based on customizable hardware configurations to accelerate light field deep learning research.

## I. INTRODUCTION

A light field (represented by L(u, v, s, t)) consists of a set of 4D light rays through every point in empty space and its intensity in a 3D scene environment, describing the amount of light flowing in every direction in space. We can represent light fields using a two-plane representation in general position (see Fig. 1) to model the analytic geometry of perspective imaging. This two-plane representation could be seen as a collection of perspective images of the ST plane each taken from an observer position on the UV plane. In another way, this can be interpreted as many cameras taking photos of the same scene at different perspective views; thus, light fields are technically a collection of photos taken at different angles. Similar to ordinary photographic cameras, light field cameras capture the quality and direction of light which allows the focus to be altered later. Because light fields contain more information than singular images, they have a lot of critical applications in the computer vision field, such as light field depth estimation [1], synthetic aperture photography [2], and 3-Dimensional models of objects.

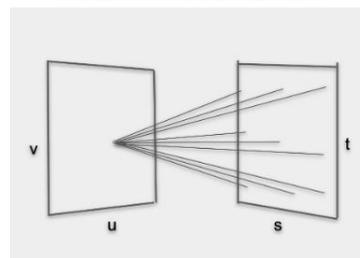

Fig. 1. *A diagrammatic representation of an individual light field, which is composed of light rays from points on the ST plane to a point on the UV plane. In another interpretation, the coordinates (u, v) describe a location of a camera on the UV plane, and the coordinates (s, t) describe the location on the ST plane through which the ray from the camera passes. The parameterization and depiction of light fields with this 2-plane representation has been adopted from Levoy and Hanrahan [3] [4].*

In the application of depth estimation, the objective is to calculate the depths of all objects in an image, or, in our case, a light field. To compute the depth of light field scenes, several manual optimization-based techniques have been used: using epipolar plane images that contain lines of different slopes, combining defocus and correspondence cues [5], etc. However, these methods either take too much time or result in low accuracies due to noise or occlusions.

In the application of synthetic aperture imaging, the goal is to measure the sharpness of a synthetic aperture image. Several calculation and optimization-based approaches for this type of application include using derivatives or local statistics of image pixel value, variance [6], discrete cosine transform [7], Tian and Chen's Laplacian mixture model of wavelets method [8], etc. However, similar to the methods in the application of light field depth estimation, these optimization approaches are either computationally expensive, inaccurate, and/or time-consuming.

Traditionally, a plenoptic camera – commonly known as a light field camera – has been used in all three light field tasks mentioned above. This camera captures light fields in a scene where many images are taken at different angles and allows photographers to alternate the focus and perspective of an image after it is captured because of the angular information of the light captured by the light field camera. However, there are limitations to its use: due to noise prevalence in most real-world light field data, depth maps, for example, which are derived from a light field captured through a plenoptic camera, appear unclear.

## II. RECENT RELATED WORK

Fortunately, recent machine learning approaches for light field applications have shown the ability to be more precise and faster than previous conventional methods, even in the presence of noise. However, there is a lack of data surrounding light fields, both in terms of the number of datasets publicly available and the amount of data in current sparse light field datasets. This motivates the creation of more robust light field datasets to fully unlock the ability of deep learning techniques, as higher quality, more varied, and larger amounts of data lead to more accurate models.

Researchers have turned to machine learning algorithms to replace past manual calculation and human-reliant approaches to solve various problems. For example, scientists used to manually sort through large amounts of data, but with a machine learning model, scientists can solve tasks such as image classification or object detection faster and more accurately. Therefore, in light field applications, machine learning algorithms have recently been more widely used. For example, in 2019, Pei et al. developed a deep neural network to estimate whether a single synthetic aperture image is in focus for the task of synthetic aperture imaging, a technique where multiple viewpoints of light fields are used to simulate a large aperture camera with a large virtual convex lens with a camera array when their images are tied together [2]. Traditionally, CNNs, due to their robust ability to learn visual features from pictures, are trained on image sets, most notably ImageNet [9], to perform image identification and classification tasks; however, more recently, CNNs have also been used to estimate depths of light fields from light field datasets, as these machine-dependent algorithms can achieve faster and more accurate light field depth estimation results than traditional methods, such as correspondence matching between views and depth from defocus with synthetic aperture photography. For example, Shin et al. [1] developed a depth estimation CNN called EPINET that achieved top rank in the HCI 4D Light Field Benchmark [10] on assessment metrics such as bad pixel ratio, mean square error, etc. The EPINET design takes in four processing streams from four directions (vertical, horizontal, right diagonal, left diagonal) of sub-aperture picture as input and outputs four independent light field representations. Then, Shin et al. combined these feature maps to produce singular and higher-level representations to estimate depth. However, the HCI synthetic light field dataset only has 28 scenes in total and the EPINET's network architecture has a small receptive field, meaning it can only handle a limited spacing between cameras, making its state-of-the-art (SOTA) performance limited and unrepresentative of performance on real-world light field data [11].

To address the lack of data surrounding light fields, Shin et al. [1] also used data augmentations, or modifications on some parts of data. Then, the modified data are added to an existing dataset of the same medium to expand the dataset. Because there is not enough variety of publicly available light field data and machine learning algorithms are usually highly dependent on the amount and variety of data to be trained accurately, Shin et al. utilized augmentations such as scaling, rotations, transpositions, etc. to increase the amount of data for sparse existing light field datasets on the HCI dataset. In the end, using augmentations helped increase the depth estimation accuracy of their EPINET algorithm. However, augmentations result in a limited increase of data, as only different crops, rotations, and other modifications of clones of the same scenes are added to an existing dataset. Hence, this process leads to a limited increase of accuracy,

Others attempted dataset creation from scratch to increase the amount and diversity of light field data. Xiong et al. [12] captured three discrete 5D hyperspectral light field scenes (represented by $f(x,y,u,v,\lambda)$) with a special hyperspectral camera. Unfortunately, this dataset is limited in size and does not have disparity labels; therefore, it cannot be used for evaluating the performance of deep learning algorithms. In addition, it takes 16 hours for their MATLAB code to reconstruct a single hyperspectral light field, and although their hybrid imager hardware system, including a Lytro camera and a coded aperture snapshot spectral imager, recovers 5D light fields with both high spectral and angular resolutions, their expensive and time-consuming process deems it impractical for others to reproduce. On the other hand, Schamback et al. [13] created a 507-scene multispectral light field dataset, where each light field is represented by L (u, v, s, t, $\lambda$), using a designed scene generator to randomly output images as well as adding seven handcrafted scenes. Though their dataset is larger than Xiong et al.'s hyperspectral light field dataset, their multispectral dataset creation approach is complex, especially having to manually handcraft seven of the scenes. Besides, most cameras, and especially most light field cameras, are not multispectral, narrowing the practical applications of using a multispectral dataset.

## III. MATERIALS AND METHODOLOGY

In this paper, we present a novel approach to create more robust light field datasets to avoid the shortcomings present in previous augmentation and creation methods and to assist learning-based methods, such as the ones mentioned above, as they are shown to outperform manual optimization-based methods for calculating light field tasks. We plan to generate different RGB (red, blue, and green) proportions of synthetic light field datasets from scratch. Our approach is easy to create, set up, and modify using a Unity engine and covers a wide range of different hardware and camera parameters that can easily be changed for any light field tasks and applications, such as the three mentioned above. In addition, the datasets created by our approach are in the RGB color format, which is significantly easier to generate and covers a wider range of applications than hyperspectral and multispectral versions, as the images captured by most cameras, including mobile, hand-held, and most significantly, plenoptic or light field cameras, are in RGB format. Furthermore, our dataset's light field images can be generated into any file type, e.g., png, jpg, etc. when being captured just by changing a method and String format in the C#

Script, which means they are scalable for many different tasks and data requirements. Our goal is to offer a practical, convenient, and robust solution to deep learning and light field researchers by creating large datasets that are reliable as well as easily scalable for numerous light field applications and a wide range of deep learning algorithms. Therefore, in this paper, we present a Unity-based approach towards increasing the diversity of light field datasets to enable more machine learning approaches that are both speed- and accuracy-efficient for a variety of applications.

First, we discuss the software and materials used to generate our datasets. To create custom light field datasets in 3D, we utilize the capabilities of Unity (see Fig. 2), a cross-platform game engine that provides a variety of virtual scenes, backgrounds, prefabs, assets, and objects. With this engine, we conveniently set up a full scene with directional lighting and terrain as the background for our light field images. Afterwards, we develop the code to tune specific hardware parameters, spawn random objects into the empty terrain space, create and position multiple cameras to view this space, and take automatic snapshots of the entire scene. The code is written in the form of C# scripts that are embedded within the Unity project in Unity Hub 3.2.0. and can be found in our public GitHub repository here [14]. We use the Unity Editor Version of 2020.3.36f1 to leverage the objects, functions, etc. that exist in and are compatible with older Unity versions. In addition, we have stored six of our datasets in .png format of various sizes (18, 18, 40, 100, 200, 500 images) generated by our Unity approach into a public Kaggle dataset for others to use [15]. We choose Kaggle because it is one of the top platforms and Jupyter notebook environments for deep learning researchers and enthusiasts alike to utilize datasets to develop machine learning algorithms. Through Kaggle, our datasets can be easily used in the same platform, downloaded to a computer, or be loaded into any other IDE (code development tool) or Jupyter notebook for training deep learning models.

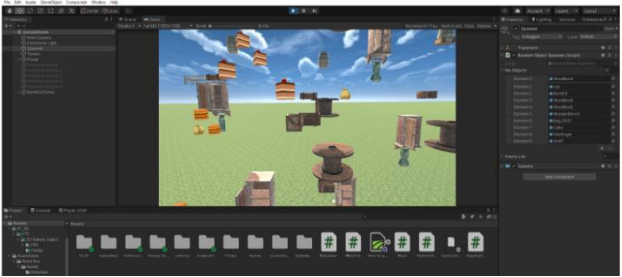

Fig. 2. *Our Unity environment set-up with objects generated. On the left are cameras and objects, in the middle is our virtual scene, and on the right are adjustable parameters of positions, scales, etc. of objects and cameras.*

Next, we discuss the process to create our custom Unity dataset. A 3D graphics pipeline (see the top sequence diagram in Fig. 3), which includes view transforms, is the starting point we use to create our light field dataset [16] [17] (Nanyang Technological University, 2012; Wetzstein). We specifically modify the Vertex Processor, one part of the graphics pipeline described above, to create light fields (see the bottom sequence in Fig. 3). We adjust both the model and view transforms of this step. We use random model transforms to create randomized scenes for an infinitely large dataset of scenes, setting up fields of view and the spawning of random objects. We create functions in our C# script to automatically generate random objects of different shapes, sizes, and textures at different positions. Additionally, we use multiple view transforms to generate a light field given a particular scene by modifying the positions of and adding multiple cameras in our Unity engine setup so they can take snapshots of the scene from multiple angles. We create functions in our C# scripts that set up a specified number of cameras at different positions and take automatic snapshots of the scene, which we can modify easily by adding different textures, objects, and backgrounds. Combining the model and view transforms, we develop an automated Unity process that can create a full light field dataset within minutes.

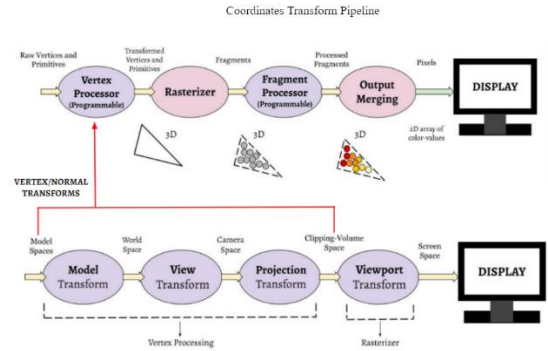

Fig. 3. *Diagram of the general 3D graphics rendering pipeline, including steps from the Vertex Processor to the 2D Display of the images, indicated by the top sequence. The bottom section indicates the process of the Vertex Processor, which translates raw vertices and primitives of objects into the vertex locations and perspectives of the viewer. This illustration for the graphics pipeline is based on the notes from Wetzstein [17].*

With the end goal of generating more light field data to train robust machine learning models, we design an end-to-end algorithm using Unity graphics and virtual scenes to generate a dataset of random images and depths of random scenes relative to any number of cameras. As mentioned above, our algorithm generates the random positioning of objects in the Unity scene and allows for adding multiple cameras for multiple view transforms to capture a full light field. By taking quick snapshots from slightly different angles, we can create a large dataset containing light field snapshots to train a deep learning architecture. We write C# scripts to automatically generate objects at random positions within set ranges and boundaries that fit the rectangular scene and to automate snapshots of these scenes. Next, the cameras capture a snapshot with a specified Width x Height resolution (can be modified in our Unity program) from each sampled location on the UV plane, delete all the objects from the scene, then regenerate new random objects at new random positions, take a snapshot of this new scene, and the entire process repeats again. At the end of the process, all the snapshots are automatically stored into a Snapshots folder.

Our Unity-based method is much faster, more convenient, scalable, reproducible, and robust than all previous methods for the reasons below. The speed of data generation is only limited by the processing speed of the computer running our Unity engine. Using a Windows 10 desktop with a 3.60GHZ CPU and 1080 Ti GPU (more details of our computer hardware specifications are listed in Fig. 5), spawning up to one thousand objects takes only about four seconds or less, and creating a 2,000 light field image dataset with 40 objects in each snapshot takes less than five and a half minutes due to our automated random object spawner script that can generate new data synthetically from any angle in the scene and take a snapshot of each light field image or scene. We also have the flexibility of creating a dataset of any size for any situation by modifying the number of images variable in Unity. We can easily generate very large light field datasets (up to 2,000 snapshots within our RAM limits) by just adjusting several parameter variables, such as producing more than a thousand different scenes for a dataset, outperforming all three datasets mentioned above- the 28-scene HCI dataset [10], 5D hyperspectral dataset [12], and 507-scene multispectral dataset [13] in size and conveniency. In addition, we can easily increase the number of viewpoints by simply updating parameters in the same Unity scene to match any new hardware, such as another plenoptic camera. Our Unity set-up also allows for the tuning and modifications of numerous parameters, including the set number of cameras, number of objects, types of objects, textures of objects, etc. to increase the diversity of produced data. Also, because we have different random scenes for each data point, we ensure no point of similarity between any data points, therefore outperforming augmentation-based methods and Schamback et al.'s two-camera system for their dataset in terms of data variety. As mentioned before, data augmentations only utilize different modifications on clones of the same scene on a dataset and are therefore not as diverse as our approach. In addition, we choose to generate RGB light fields, where their three-channel colors replace light fields' coordinates and spectral dependencies [13], because RGB light fields are more easily generated and are more widely applicable for light field tasks than multispectral light fields, as most cameras use RGB formats. Therefore, our approach also outperforms Schamback et al.'s multispectral dataset by a second assessment metric [13]. Furthermore, our Unity-based approach is flexible, as its data generation parameters can easily be changed for different hardware setups and matches exactly with specific hardware parameter requirements, such as cameras' resolutions and positions. For Xiong et al.'s hyperspectral dataset, they included different light field images, but their datasets were only made for a specific hardware setup. To make their datasets suitable for different hardware setups, their datasets must be changed or regenerated to match that specific hardware setup, which is cumbersome as they need to use both a Lytro camera and an imager to recover light fields. For our method, the parameters of our snapshots and locations on the UV plane can be updated easily in Unity to match specified hardware parameters, such as the number and position of cameras, etc. Our Unity scenes simulate the occlusions, reflections, and diffractions of light present in real-world scenes, which allows us to quickly generate accurate light field datasets without noise, without expensive hardware setups such as plenoptic cameras or spectral imagers, or human intervention. The only materials one would need to generate a light field dataset are a Unity3D engine and Unity Hub/Editor installed on a desktop or laptop and our publicly available code, which takes just minutes to create a light field image dataset of any size. Most notably, we developed our C# script to automate the generation of objects and capturing of images, requiring no manual work other than changing variables and parameters. Therefore, our approach is affordable, customizable, convenient, and reproducible.

## IV. RESULTS

Our Unity program outputs a set number of images with the cameras at different x and y coordinate positions. As discussed in the previous section, the C# scripts that indicate variable parameters can be customized to change the number of images to capture, the number of translations the cameras make, and the distance the cameras move. As well as the viewpoint of the scene, every parameter within the scene can be customized to fit specific simulation specifications. Fig. 4 displays one of our datasets.

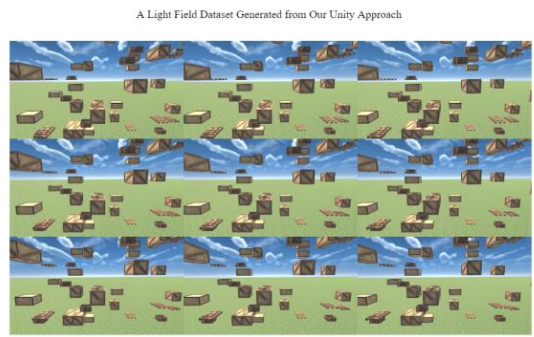

Fig. 4. *Different x and y coordinate views of randomly generated objects. From left to right the cameras were translated right and from top to bottom they were translated down.*

The amount, size, and quantity of objects spawned can be adjusted to simulate real-life measurements and quantities and are only limited by hardware (our specific computer hardware parameters are listed in the caption of Fig. 5). Our program is lightweight, computationally efficient, and CPU-efficient where a substantial number of objects can be spawned at once. Up to 1,000 generated objects in a single scene are tested and proven to be fast and stable with a variety of Unity hardware and variable parameter specifications, and up to 2,000 light field images for a dataset can be generated within five and a half minutes (see Fig. 5 for a comparison of data creation times).

DATASET CREATION SPEED USING OUR APPROACH

| Dataset Size, in number of snapshots | Dataset Creation Time, in minutes and seconds |
|---|---|
| 100 | 16 seconds |
| 200 | 31 seconds |
| 500 | 79 seconds (1 minute 19 seconds) |
| 1,000 | 159 seconds (2 minutes 39 seconds) |
| 2,000 | 325 seconds (5 minutes 25 seconds) |

Fig. 5. *Table showing the speed of our dataset generation approach based on several dataset sizes. Each snapshot of a scene has 40 objects. The quick speed ensures the reproducibility and convenience of our method in generating new light field datasets, which is much faster than Xiong et. al's method, which took 16 hours. Our Unity program was run on a Windows 10 desktop, Version 10.0.19044 and Build 19044 with a Intel Core i9-9900K @ 3.60GHZ CPU, NVIDIA GeForce GTX 1080 Ti GPU, and 32.0 GB @ 3200 MHz RAM.*

## V. CONCLUSION

In this paper, we presented a novel approach for generating customizable light field image datasets that are quick, easy, customizable, and robust for machine learning. Deep learning-based architectures, such as EPINET [1], have been shown to be more robust, accurate, and faster than light field cameras and optimization-based techniques for the above tasks. Overall, our custom curated dataset can be used for machine learning models in an expansive variety of light field applications, notably synthetic aperture photography, depth estimation, 3D representations, and more.

## VI. LIMITATIONS AND FUTURE DIRECTIONS

Because Unity provides virtual scenes only, all our datasets are synthetic and without noise, which is unrepresentative of the real world as real-life cameras and images may include noise. Hence, in the future, we plan to simulate real-world noise by adding Unity3D's built in Cinemachine Noise Properties, such as Perlin noise, which helps simulate random movements to our virtual cameras [18]. Thanks to the promise of machine learning as a robust and accurate mechanism for numerous applications, specifically for light field tasks, our novel Unity-based approach can easily be used as a reproducible mechanism to automatically create large, variegated datasets to train any machine learning model to maximize its efficiency and accuracy. For future research, we plan to develop a novel machine learning model ourselves, such as a convolutional neural network or autoencoder, to test the robustness and usability of our Unity-generated light field datasets.

## ACKNOWLEDGMENT

We would like to thank the SHTEM Summer Research Internship Program managed by the Stanford University Compression Forum and its organizers Cindy Nguyen, Sylvia Chin, Carrie Lei, Eric Guo, and Kaiser Williams for providing us amazing opportunities to participate in scientific research. Specifically, we would like to thank our mentor Manu Gopakumar, an Electrical Engineering PhD student from Stanford University, for his guidance throughout our research.